\documentclass{article}
\usepackage{spconf,amsmath,graphicx}
\usepackage{amsfonts}
\usepackage{booktabs, makecell, multirow, adjustbox}
\usepackage{flushend}
\usepackage{bm}
\usepackage{enumitem}
\usepackage{cite}
\usepackage{algorithm}
\usepackage{algorithmic}
\usepackage{hyperref}

\usepackage[capitalize,noabbrev]{cleveref}
\usepackage{xcolor}

\def\method{ETAGE}
\def\bS{\mathbb{S}}
\def\bT{\mathbb{T}}

\title{ETAGE: ENHANCED TEST TIME ADAPTATION WITH INTEGRATED ENTROPY AND GRADIENT NORMS FOR ROBUST MODEL PERFORMANCE}
\name{Afshar Shamsi$^{\dag}$, Rejisa Becirovic$^{\dag\dag}$, Ahmadreza Argha$^{\dag\dag}$,
Ehsan Abbasnejad$^{\ddag}$\vspace{-.14in}} 
\address{\textit{Hamid Alinejad-Rokny$^{\dag\dag}$, Arash Mohammadi$^{\dag}$\thanks{This work was partially supported by the Natural Sciences and Engineering Research Council (NSERC) of Canada through the NSERC Discovery Grant RGPIN-2023-05654}}\\\\
$^{\dag}$ Concordia Institute of Information Systems Engineering, Concordia University, Montreal, Canada \\ $^{\ddag}$ Australian Institute for Machine Learning (AIML), University of Adelaide, Australia \\ $^{\dag\dag}$ School of Biomedical Engineering, UNSW Sydney, Sydney, Australia}
\begin{document}
\ninept
\maketitle
\begin{abstract}
\vspace{-2mm}
Test time adaptation (TTA) equips deep learning models to handle unseen test data that deviates from the training distribution, even when source data is inaccessible. While traditional TTA methods often rely on entropy as a confidence metric, its effectiveness can be limited, particularly in biased scenarios. Extending existing approaches like the Pseudo Label Probability Difference (PLPD), we introduce ETAGE, a refined TTA method that integrates entropy minimization with gradient norms and PLPD, to enhance sample selection and adaptation. Our method prioritizes samples that are less likely to cause instability by combining high entropy with high gradient norms out of adaptation, thus avoiding the overfitting to noise often observed in previous methods. Extensive experiments on \texttt{CIFAR-10-C} and \texttt{CIFAR-100-C} datasets demonstrate that our approach outperforms existing TTA techniques, particularly in challenging and biased scenarios, leading to more robust and consistent model performance across diverse test scenarios. The codebase for ETAGE is available on~\url{https://github.com/afsharshamsi/ETAGE}.
\end{abstract}
\vspace{-.2in}
\section{Introduction} \label{sec:intro}
\vspace{-.1in}

Deep learning models~\cite{shamsi2021uncertainty, habibpour2023uncertainty, simonyan2014very} have demonstrated significant success across various tasks, particularly when both training and testing data share the same distribution. In real-world applications, however, such models often face data that deviates from training distribution~\cite{koh2021wilds, pan2009survey}, a challenge known as domain shift or dataset shift. This discrepancy between the training (source) data and the testing (target) data can severely undermine model performance~\cite{recht2018cifar}, thereby limiting their practical efficiency. Tackling this issue requires innovative strategies that enable models to adapt to new data distributions without requiring additional training data or supervision.

\vspace{.025in}
\noindent
\textbf{\textit{Literature Review:}} Unsupervised Domain Adaptation (UDA)~\cite{ganin2015unsupervised, park2020joint} has been one approach to address the aforementioned issue. Generally speaking, in the UDA approach,  the knowledge from labeled source data is transferred to unlabeled target data using both datasets during the training. By analyzing the distribution of the target set, UDA allows the model to learn domain-invariant features~\cite{pinheiro2018unsupervised} that generalize well across different distributions. While UDA has shown promise in various scenarios, it still relies on the availability of source data during the adaptation process. Test Time Adaptation (TTA)~\cite{azimi2022self} bridges this gap by enabling models to adapt solely at test time, without access to the source data. This makes TTA particularly appealing in situations where retraining is impractical due to privacy concerns, computational constraints, or the urgent need for adaptation.

Test-Time Entropy Minimization (TENT)~\cite{wang2020tent} is among the first strategies proposed for TTA, which focused on reducing the entropy of model predictions during test time. TENT enhances model robustness to corrupted datasets and domain adaptation scenarios by directly minimizing prediction entropy, i.e., updating model parameters through entropy minimization without altering the training process. Entropy-Aware Test-time Adaptation (EATA)~\cite{niu2022efficient} is another noteworthy TTA approach that extended TENT by stabilizing model weights to address the ``forgetting'' phenomenon, i.e., degradation of adapted models' performance on in-distribution test samples. EATA starts with an initial pass over the target data to estimate the distribution, applying Fisher-based~\cite{kirkpatrick2017overcoming} weighting to identify crucial parameters for adaptation. This ensures that critical model weights remain stable and minimizes the impact of noisy or irrelevant samples. Sharpness-Aware Minimization (SAR)~\cite{niu2023towards} further improves stability by optimizing only the most reliable features.  By guiding reliable samples towards a flat minimum, it enhances the model's stability and resilience, even under severe distribution shifts. SAR is based on the insight that models trained in flatter regions of the loss landscape tend to be more robust to perturbations~\cite{li2018visualizing}, thereby, ensuring consistent performance even under challenging conditions. Despite the advancements made, existing approaches to TTA focus primarily on entropy as the key measure of confidence. This focus, however, has limitations, particularly in scenarios where spurious correlation shift makes entropy an unreliable confidence metric~\cite{beery2018recognition}. Recently, DeYO~\cite{lee2024entropy} proposed the concept of Pseudo Label Probability Difference (PLPD) to better identify harmful samples that may compromise model's performance. DeYO enhances performance by intentionally degrading the shape of objects in images, ensuring the model's judgments are based on generalizable features rather than misleading patterns. 

\vspace{.025in}
\noindent
\textbf{\textit{Contributions:}} Building upon SAR's focus on mitigating noisy samples through sharpness aware minimization and DeYO's use of shape information in TTA,  we propose the enhanced test time adaptation with integrated entropy and gradient norms (ETAGE) method. ETAGE introduces a refined test-time adaptation strategy that integrates entropy minimization with gradient norms and PLPD directly addresses the limitations of SAR and DeYo. By considering gradient norms, we capture the model's sensitivity to noisy samples more effectively, enhancing the stability and efficacy of the adaptation process. Additionally, we provide a mathematical analysis demonstrating why PLPD alone may miss noisy gradient samples, and propose a filtering method to eliminate these samples by combining high entropy with high gradient norms out of adaptation. The proposed ETAGE method not only avoids overfitting to noise but also ensures more robust performance across different test sets, as supported by empirical evidence/results from our experiments. In short, the paper makes the following key contributions:
\begin{itemize}[noitemsep]
\item Introduction of a refined test-time adaptation approach, ETAGE, that couples entropy minimization with gradient norms and PLPD. ETAGE filters out noisy gradient samples by combining high entropy and high gradient norms, avoiding overfitting to noise.
\item Theoretically illustrating shortcomings of PLPD in identifying noisy gradient samples, and mathematically demonstrating how to address this limitation.
\item To the best of our knowledge, this is the first implementation of Contrastive Language-Image Pre-training (CLIP) foundation model~\cite{ilharco_gabriel_2021_5143773} for TTA.
\end{itemize}
For performance evaluations, extensive experiments were conducted on \texttt{CIFAR-10-C} and \texttt{CIFAR-100-C}. It is observed that ETAGE achieves superior generalization and more consistent performance across different test sets compared to its state-of-the-art counterparts.
The rest of the paper is organized as follows: First,~\cref{sec:background} provides background information required for presentation of the proposed TTA approach. The ETAGE method is then introduced in~\cref{sec:method}. \cref{sec:results}, first, presents the datasets utilized in this study, and then provides in details experimental results. Finally~\cref{sec:conclusiom} concludes the paper.

\setlength{\textfloatsep}{0pt}
\begin{figure}[t!]
    \begin{minipage}[b]{.3\linewidth}
      \centering
      \centerline{\includegraphics[width=2.15cm]{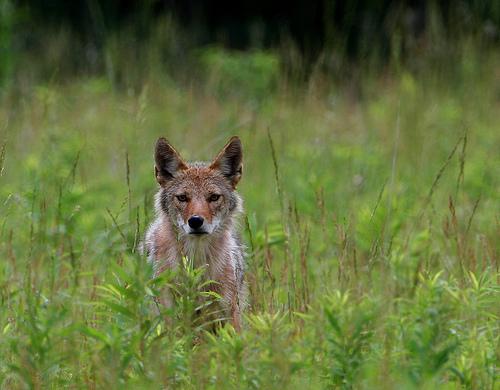}}
      \centerline{(a)  \small{Original}}\medskip
    \end{minipage}
    \begin{minipage}[b]{.3\linewidth}
      \centering
      \centerline{\includegraphics[width=2.15cm]{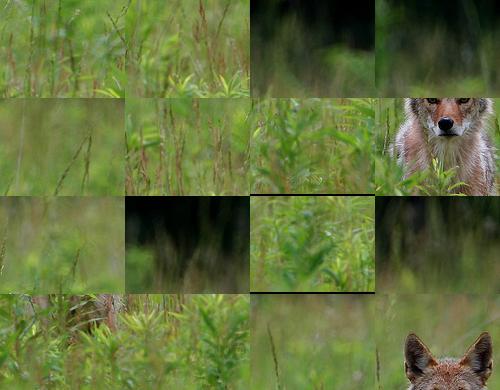}}
      \centerline{(b) \small{Patch Size 4}}\medskip
    \end{minipage}
    \begin{minipage}[b]{.3\linewidth}
      \centering
      \centerline{\includegraphics[width=2.15cm]{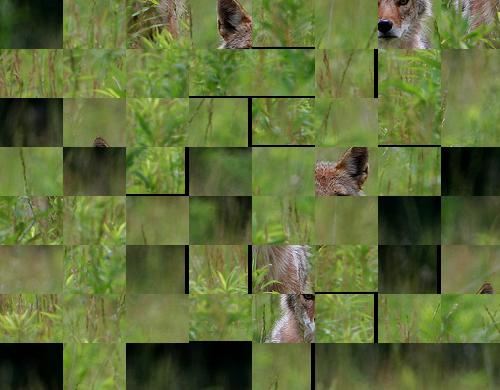}}
      \centerline{(c) \small{Patch Size 8}}\medskip
    \end{minipage}
    \vspace{-5mm}
    \caption{\footnotesize{Patch augmentation for estimating PLPD.}}
    \label{fig:res}
\end{figure}

\vspace{-.15in}
\section{PRELIMINARIES} \label{sec:background}
\vspace{-.1in}
In this section, we provide the required background used for development of the proposed ETAGE method. 

\vspace{.05in}
\noindent
\textbf{2.1. Distribution Shift}

\noindent
Domain shift refers to differences between the distributions of training and testing data, which can be formally described as follows
\begin{equation}\label{equation:source_dataset}    \nonumber
    \bS = \{ (\mathbf{x},\mathbf{y})| \forall(\mathbf{x},\mathbf{y}) \sim p_\bS\}, \quad f_\bS: f_\bS(\mathbf{x})=\mathbf{y}
\end{equation}
\begin{equation}\label{equation:target_dataset}  \nonumber
    \bT = \{ (\mathbf{x},\mathbf{y}) | \forall(\mathbf{x},\mathbf{y}) \sim p_\bT\}, \quad f_\bT: f_\bT(\mathbf{x})=\mathbf{y}
\end{equation}
where $(\mathbf{x}, \mathbf{y})$ represents the source data, which follows a probability distribution denoted by $p_\bS$. Function $f_\bS(\cdot)$ is a mapping that assigns $\mathbf{x}$ to $\mathbf{y}$. The target data, $\bT$, is defined in a similar fashion.
Consider a Neural Network (NN) denoted by $f(\mathbf{x}, \boldsymbol{\theta}): \mathbf{x} \rightarrow \mathbf{y}$, where $f(\cdot)$ represents a fixed architecture with parameters $\boldsymbol{\theta}$. Generally speaking, the objective is to minimize the loss function given by
\begin{equation}\label{equation:loss1}  
    \varepsilon_T(\boldsymbol{\theta}) = \mathbb{E}_{(\mathbf{x},\mathbf{y}) \sim P_T}[\ell(f(\mathbf{x},\boldsymbol{\theta}),\mathbf{y})].
    \nonumber
\end{equation}
Such a minimization task becomes particularly challenging in a TTA setup, where access to the source data is restricted.

\vspace{.05in}
\noindent
\textbf{2.2. Entropy Minimization}

\noindent
To tackle the above mentioned issue, where the model is required to adapt to new, unseen data during inference without access to the source data, one effective strategy is entropy minimization. The key idea behind this approach is to adjust the model's parameters in real-time, focusing on minimizing the entropy of the output predictions. Entropy, in this context, measures the uncertainty of the model's predictions. By minimizing entropy, the model becomes more confident in its predictions, effectively adapting to the new distribution presented by the test data. To achieve this, typically, the focus is on adapting only specific layers within the model, particularly normalization layers, which are believed to serve as proxies for the source data by retaining certain statistics of the source domain. This approach was first introduced with Batch Normalization (BN) layers in TENT, and has since been extended to other normalization techniques such as Group Normalization (GN) and Layer Normalization (LN)~\cite{niu2023towards}. By adjusting these normalization layers during test time, the model can adapt to the new data distribution without requiring access to the source data, therefore,  maintaining its performance across varying domains.

\vspace{.05in}
\noindent
\textbf{2.3. Pseudo Label Probability Difference (PLPD)}

\noindent
PLPD focuses on estimating the probability associated with an input for the same class before and after applying noise or augmentation, and is computed as follows
\begin{equation} \label{eq:plpd}
\text{PLPD}_{\boldsymbol{\theta}}(\mathbf{x},\mathbf{x}') = \left(\mathbf{P}_{\boldsymbol{\theta}}(\mathbf{x}) - \mathbf{P}_{\boldsymbol{\theta}}(\mathbf{x}')\right)_{\hat{y}'},
\end{equation}
where $\mathbf{x}$ and  $\mathbf{x}'$ are the sample input before and after, respectively, applying the augmentation/noise. The PLPD measures the sensitivity of the model's predictions to slight changes in the input. For example, the patch shuffle technique, as shown in \cref{fig:res}, rearranges small regions (patches) in the input to alter spatial dependence within the image. By evaluating the PLPD under such conditions, one can assess the accuracy of the model when key information is disrupted.
The core idea is that if the model still assigns the same class to the input after noise or augmentation, it may have learned spurious correlations from the source, rather than focusing on the shape of the object in the input image. The noise or augmentation is, typically, designed to disrupt or destroy objects within the image. The desired scenario is a high PLPD, which indicates that after the object is corrupted, the model is unable to assign the same class. This scenario is crucial as it identifies samples that should be used later for adaptation, ensuring the model's predictions are based on more meaningful features. 

\vspace{-.15in}
\section{The ETAGE Method}
\label{sec:method}
\vspace{-.1in}

In this section, we introduce the proposed \method, that leverages both a gradient norm threshold and PLPD to enhance sample filtering. We premise that a high gradient norm indicates that the model is highly sensitive to small perturbations in the input space. Such samples, even if they produce an acceptable PLPD, might be misleading because the underlying instability (captured by the high gradient norm) is not reflected in the PLPD calculation. To explore this premise, a mathematical counterexample is provided below. 

\begin{figure}[t!]
\begin{minipage}[b]{.48\columnwidth}
  \centering
  \includegraphics[width=3.5cm, height= 2.7cm]{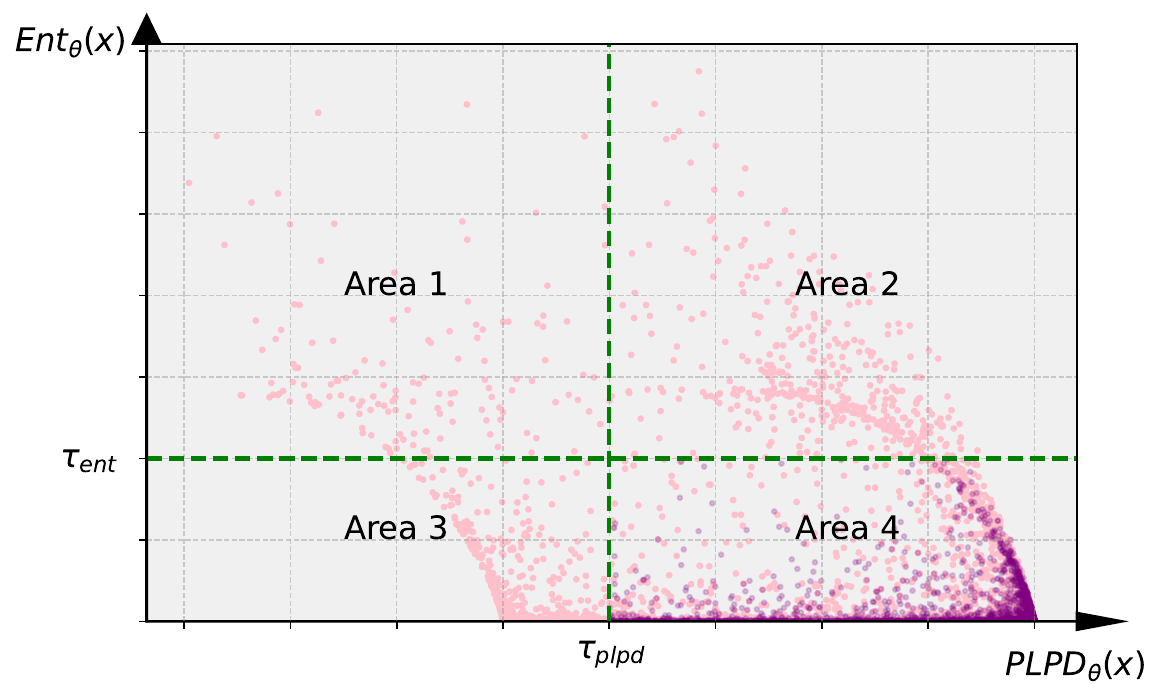}
  \centerline{(a) \small{Areas}}\medskip
  \label{fig:acc}
 \end{minipage} 
 \begin{minipage}[b]{.48\columnwidth}
  \centering
  \includegraphics[width= 3.5cm, height= 2.5cm]{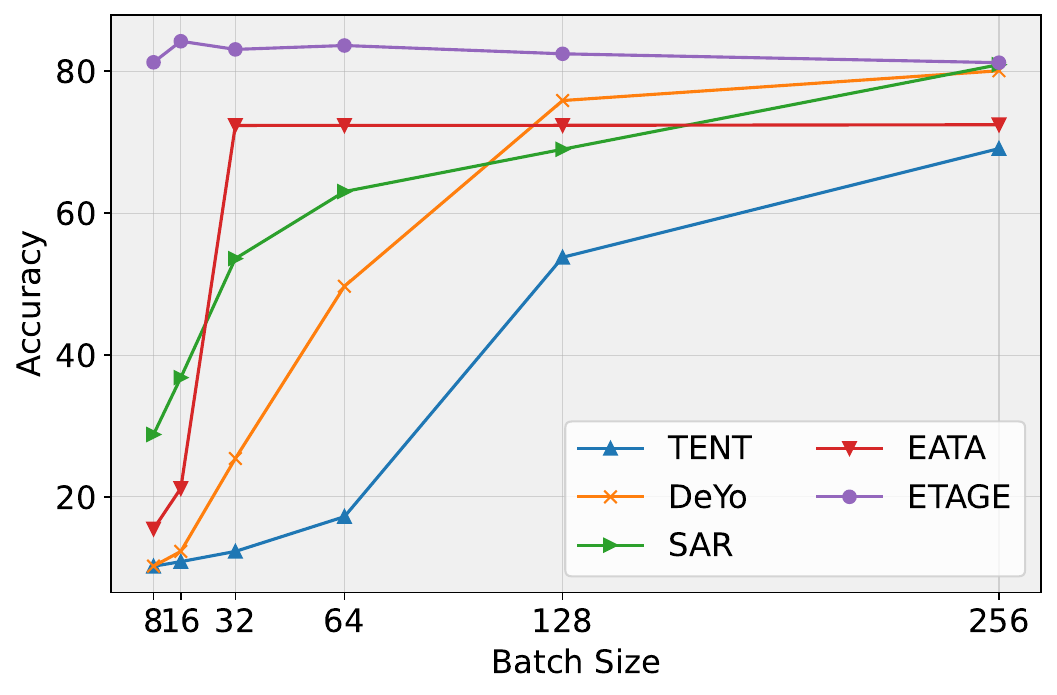}
  \centerline{(b) \small{Batch size versus accuracy}}\medskip
  \label{fig:area}
 \end{minipage} 
\vspace{-4mm}
\caption{\footnotesize{The left figure classifies the test set data into four different areas using three thresholds: entropy, gradient norm, and PLPD. It should be noted that only the points/samples in purple, located in area $4$, pass the thresholds and will be used for \method. The right figure shows the performance of different methods under various batch sizes for Gaussian noise on \texttt{CIFAR-10-C}.}}
\label{fig:batch}
\vspace{0.025in}
\end{figure}

\begin{algorithm}[!t]
\caption{Online Adaptation for \method}
\label{alg:tta}
\begin{algorithmic}[1] 
    \STATE \textbf{Input:} Model, Test samples
    \STATE \textbf{Output:} Adapted Model
        \FOR{each batch}
            \STATE Compute norm, and entropies
            \STATE Applying entropy and norm thresholds
            \IF{filtered samples exist}
                \STATE Patch permutation on remaining samples
                \STATE Calculate PLPD
                \STATE Filter samples based on PLPD threshold
                \IF{filtered samples exist}
                    \STATE Compute loss using remaining entropies
                    \STATE Backpropagate loss to update model
                    \STATE Update optimizer for the model
                \ENDIF
            \ENDIF
        \ENDFOR
\end{algorithmic}
\end{algorithm}
ETAGE, first applies a gradient norm threshold in addition to an entropy threshold to ensure that only stable samples proceed to the PLPD calculation. The gradient norm is used to identify samples that may cause instability in the model's predictions, and is defined~as
\begin{equation}\label{eq:norm}
\|\nabla_{\boldsymbol{\theta}} \mathcal{L}(\mathbf{x}, {\boldsymbol{\theta}})\| \propto \left\|\frac{\partial P(\mathbf{y} \mid \mathbf{x},{\boldsymbol{\theta}})}{\partial \mathbf{x}}\right\|.
\end{equation}
A high gradient norm indicates that the model's output is highly sensitive to small changes in the input, which can be a sign of overfitting to noise or instability in the learned decision boundary. Now, consider a sample \( x \) where the gradient norm is very high, i.e.,
\begin{eqnarray}
\|\nabla_{\mathbf{x}} \mathcal{L}(\mathbf{x}, {\boldsymbol{\theta}})\|_2 \gg \tau_{\text{Grad}},\nonumber
\end{eqnarray}
where $\tau_{\text{Grad}}$ is the gradient norm threshold. A high gradient norm suggests that even a small change in \( \mathbf{x} \) will cause a significant change in the model's prediction. Suppose we apply a small perturbation \( \delta \) to \( \mathbf{x} \), resulting in \( \mathbf{x'} = \mathbf{x} + \delta \). Using the first-order Taylor expansion, the predicted probability is approximated as
\begin{eqnarray}
P(\mathbf{y} \mid \mathbf{x'},{\boldsymbol{\theta}}) \approx P(\mathbf{y} \mid \mathbf{x}, {\boldsymbol{\theta}}) + \nabla_{\mathbf{x}} P(\mathbf{y} \mid \mathbf{x}, {\boldsymbol{\theta}}) \cdot {\delta}.
\end{eqnarray}
\cref{eq:plpd} can then be approximated as
\begin{align}\nonumber
\text{PLPD}_{\boldsymbol{\theta}}(\mathbf{x}, \mathbf{x'}) &= \left( P(\mathbf{y} \mid \mathbf{x},{\boldsymbol{\theta}}) - \left( P(\mathbf{y} \mid \mathbf{x}, {\boldsymbol{\theta}}) \right. \right. \\
&\quad \left.\left. + \nabla_{\mathbf{x}} P(\mathbf{y} \mid \mathbf{x}, {\boldsymbol{\theta}}) \cdot {\delta} \right) \right)_{\hat{\mathbf{y}}},
\end{align}
which simplifies to
\begin{equation}
\text{PLPD}_{\boldsymbol{\theta}}(\mathbf{x}, \mathbf{x'}) \approx -\left( \nabla_{\mathbf{x}} P(\mathbf{y} \mid \mathbf{x}, {\boldsymbol{\theta}}) \cdot {\delta} \right)_{\hat{\mathbf{y}}}.
\end{equation}
The value of PLPD can be further estimated as
\begin{equation}
   \text{PLPD}_{\boldsymbol{\theta}}(\mathbf{x}, \mathbf{x'}) \approx \|\nabla_{\mathbf{x}} P(\mathbf{y} \mid \mathbf{x}, {\boldsymbol{\theta}})\|_2 \|{\delta}\|_2 \cos({\boldsymbol{\theta}}),
\end{equation}
where \( {\boldsymbol{\theta}} \) is the angle between \( \nabla_{\mathbf{x}} P(\mathbf{y} \mid \mathbf{x}, {\boldsymbol{\theta}}) \) and \( \delta \). The PLPD values primarily depend on two components: the gradient norm \( \|\nabla_{\mathbf{x}} P(\mathbf{y} \mid \mathbf{x}, {\boldsymbol{\theta}})\|_2 \) and the perturbation magnitude \( \|{\delta}\|_2 \). Now, consider a scenario where the input is highly sensitive to small perturbations. In such cases, as illustrated in \cref{eq:norm}, these samples will generate high gradient norms, meaning the PLPD is largely influenced by \( \|\nabla_{\mathbf{x}} P(\mathbf{y} \mid \mathbf{x},  {\boldsymbol{\theta}})\|_2 \). Consequently, even with low perturbations, the PLPD values will be disproportionately high. This undermines the reliability of the PLPD since it is meant to reflect the difference in predicted probabilities before and after perturbation. In other words, for noisy gradient samples, the reason the PLPD is high (and why such samples might pass the threshold) is due to the high gradient norm. To tackle this issue, we refine the sample selection to exclude those with high gradient norms, i.e.,
\begin{align}
    \mathbf{S'}_{\boldsymbol{\theta}}(\mathbf{x}) = \{ \mathbf{x} \mid &\ \text{Ent}_{\boldsymbol{\theta}}(\mathbf{x}) > \nonumber\tau_{\text{Ent}}, 
   \ \|\nabla_{\boldsymbol{\theta}} \mathcal{L}(\mathbf{x}, {\boldsymbol{\theta}})\| < \tau_{\text{Grad}}\\ &,\text{PLPD}_{\boldsymbol{\theta}}(\mathbf{x}, \mathbf{x'}) > \tau_{\text{PLPD}} \} \nonumber
\end{align}
We remove these noisy, high-gradient samples from the set, leading to a more stable and effective adaptation process. The final entropy loss to be minimized is given by
\begin{equation}
    \mathcal{L}_{\text{final}}({\boldsymbol{\theta}}) = \sum_{\mathbf{x} \in \mathbf{S'}_{\boldsymbol{\theta}}(\mathbf{x})} \text{Ent}_{\boldsymbol{{\boldsymbol{\theta}}}}(\mathbf{x}).
\end{equation}
\cref{alg:tta} outlines steps to perform ETAGE method. \cref{fig:batch} also presents different samples of \texttt{CIFAR-10-C} for Gaussian noise distortion based on the entropy and PLPD generated by the CLIP model. Following the common approach~\cite{wang2020tent},  a threshold, $\tau_{ent}$, is defined for entropy selection to remove samples with low confidence (Areas $1$ and $2$ in \cref{fig:batch}(a)). For the remaining samples, then the norm of the gradients are estimated to remove noisy samples that will harm the adaptation later on. Assessing the PLPD and keeping those samples with high PLPD (larger than the threshold) is the next step performed to remove samples that model tries to predict without considering the object in the image (i.e., model memorized such from the source domain). 

 As it is depicted in Area $4$ of~\cref{fig:batch,}(a), ETAGE further refines the sample selection process for adaptation by removing noisy gradient samples and only utilizing healthy samples (those shown in purple). This in turn leads to adapting the model with fewer samples. 
The proposed ETAGE method differs from SAR~\cite{niu2023towards} in the sense that the latter fails to explicitly identify and remove noisy gradient samples. Different from our direct approach, SAR only indirectly mitigates the effects of noisy gradient samples using sharpness aware minimization. ETAGE also differs from DeYO~\cite{lee2024entropy} in the sense that the latter does not take into account the harmful effect of noisy samples with high gradient norm that can significantly degrade the adaptation performance.


\vspace{-.15in}
\section{Simulations and results}\label{sec:results}
\vspace{-.1in}
\begin{table*}[!t]
    \centering
    \caption{\footnotesize{Performance of different methods on various corruptions, each with severity of $5$, on \texttt{CIFAR-10-C} dataset regarding accuracy ($\%$)}. The best-performing results are in \textbf{bold}, the second-best in \underline{underline}.}
    \label{tab: cifar10}
    \begin{adjustbox}{width=\textwidth}
    \begin{tabular}{lcccccccccccc}
        \toprule
        \multirow{2}{*}{Method} & \multicolumn{12}{c}{\texttt{CIFAR-10-C}} \\
        \cmidrule(lr){2-13}
         & Brightness & Contrast & \makecell{Elastic\\Transform} & \makecell{Gaussian\\Blur} & \makecell{Gaussian\\Noise} & \makecell{JPEG\\Compression} & Pixelate & Saturate & \makecell{Shot\\Noise} & Spatter & \makecell{Speckle\\Noise} & Average \\
        \midrule
        TENT & \underline{96.61} & \textbf{95.84} & 89.65 & 93.86 & 53.78 & 36.06 & 93.22 & 94.39 & 55.00 & 92.60 & 60.19 & 78.29 \\
        DEYO & 94.25 & 94.70 & 87.47 & 91.45 & \underline{75.88} & 79.41 & 91.76 & 93.24 & 81.39 & 91.16 & 81.17 & 87.44 \\
        SAR & \textbf{96.65} & \textbf{95.84} & \textbf{91.50} & \underline{94.51} & 69.01 & \underline{86.00} & \textbf{94.26} & \underline{95.28} & \underline{84.41} & 94.29 & 79.06 & \underline{89.16} \\
        EATA & 96.58 & 91.78 & 91.14 & 91.38 & 72.38 & 84.76 & 86.66 & \textbf{95.35} & 79.00 & \textbf{94.78} & \underline{81.35} & 87.74 \\
        ETAGE & 96.18 & \underline{95.65} & \underline{91.34} & \textbf{94.52} & \textbf{82.47} & \textbf{86.45} & \underline{93.59} & 95.17 & \textbf{86.08} & \underline{94.38} & \textbf{86.68} & \textbf{91.14} \\
        \bottomrule
    \end{tabular}
    \end{adjustbox}
    \vspace{-.25in}
\end{table*}
\begin{table*}[!t]
    \centering
    \caption{\footnotesize{Performance of different methods on various corruptions, each with severity of $5$, on \texttt{CIFAR-100-C} dataset regarding accuracy ($\%$)}. The best-performing results are in \textbf{bold}, the second-best in \underline{underline}.}
    \label{tab: cifar100}
    \begin{adjustbox}{width=\textwidth}
    \begin{tabular}{lccccccccccccc}
        \toprule
        \multirow{2}{*}{Method} & \multicolumn{12}{c}{\texttt{CIFAR-100-C}} \\
        \cmidrule(lr){2-13}
         & Brightness & Contrast & \makecell{Elastic\\Transform} & \makecell{Gaussian\\Blur} & \makecell{Gaussian\\Noise} & \makecell{JPEG\\Compression} & Pixelate & Saturate & \makecell{Shot\\Noise} & Spatter & \makecell{Speckle\\Noise} & Average \\
        \midrule
        TENT & 84.31 & 81.61 & 61.93 & 77.09 & 16.19 & 23.94 & 77.79 & 77.88 & 32.97 & 77.22 & 22.12 & 57.55 \\
        DEYO & 84.89 & 83.03 & 73.64 & 79.94 & \underline{57.24} & 64.65 & 79.65 & 80.01 & \textbf{65.63} & 79.07 & \textbf{66.86} & \underline{74.06} \\
        SAR & \textbf{85.76} & \underline{83.28} & \textbf{74.90} & \textbf{80.91} & 41.28 & \textbf{66.32} & \textbf{80.14} & \textbf{80.72} & 51.46 & \textbf{79.69} & 64.75 & 71.75 \\
        EATA & 1.70 & 1.33 & 1.56 & 1.49 & 1.37 & 1.47 & 1.51 & 1.53 & 1.40 & 1.64 & 1.26 & 1.48 \\
        ETAGE & \underline{85.18} & \textbf{83.43} & \underline{74.14} & \underline{80.36} & \textbf{61.03} & \underline{65.75} & \underline{79.72} & \underline{80.56} & \underline{64.94} & \underline{79.14} & \underline{66.17} & \textbf{74.58} \\
        \bottomrule
    \end{tabular}
    \end{adjustbox}
    \vspace{-3mm}
\end{table*}

To evaluate performance of the ETAGE method, we have used CLIP~\cite{ilharco_gabriel_2021_5143773}, which is a large pre-trained model that utilizes contrastive learning to connect image and text data. This approach was first introduced in the groundbreaking CLIP research~\cite{Radford2021LearningTV}. CLIP is composed of two main components: a text encoder and an image encoder. The text encoder is based on a Transformer architecture, which effectively processes and understands natural language input. The image encoder, on the other hand, can be implemented using either a Vision Transformer (ViT) (the one that we have utilized is the ViT-B/32) or a Convolutional Neural Network (CNN). Together, these encoders are trained to align images with their corresponding text descriptions, enabling the model to perform various tasks involving both visual and textual information. For the CLIP model we modify LN layers while keeping the rest parameters freezed in the adaptation phase. 

\begin{table}[t!]
    \centering
    \caption{\small{comparison of calibration and discriminative metrics across different methods on \texttt{CIFAR-10-C}}.}
    \label{tab:metrics}
    \begin{tabular}{lcccc}
        \toprule
        Method & ECE & MCE & Brier & AUROC \\
        \midrule
        TENT & 0.4389 & 0.4539 & 0.08972 & 0.7905 \\
        DEYO & 0.1945 & 0.4144 & \underline{0.04251} & 0.9618 \\
        SAR & 0.2731 & \textbf{0.3699} & 0.05783 & 0.8861 \\
        EATA & \underline{0.1888} & 0.3981 & 0.04528 & \underline{0.9633} \\
        ETAGE & \textbf{0.1327} & \underline{0.3727} & \textbf{0.03007} & \textbf{0.9793} \\
        \bottomrule
    \end{tabular}
\end{table}
\vspace{-.15in}
\subsection{Datasets}\label{sec:dataset}
\vspace{-.1in}
We employed CIFAR-10-C and CIFAR-100-C datasets in our experiments to evaluate the ETAGE method. Below, we provide a brief yet comprehensive introduction to these datasets, highlighting their key characteristics and relevance to our research objectives.

\vspace{-.15in}
\subsubsection{\texttt{CIFAR-10-C}}
\vspace{-.075in}
This dataset shares its training samples with \texttt{CIFAR-10}~\cite{krizhevsky2009learning}, but the test set, \texttt{CIFAR-10-C}~\cite{doan2024bayesian}, is created by applying various distortions to the original \texttt{CIFAR-10} test images. These modifications include a range of corruptions and noises, such as blur, Gaussian noise, and shot noise. \texttt{CIFAR-10-C} presents these corruptions at five different levels of severity, resulting in a total of $50,000$ test images for each type of noise. This dataset is particularly valuable for assessing the performance of image recognition models across diverse real-world scenarios, therefore, contributing to the enhancement of their reliability and robustness.

\vspace{-.15in}
\subsubsection{\texttt{CIFAR-100-C}}
\vspace{-.075in}

This dataset extends \texttt{CIFAR-100} by applying a range of corruptions (similar to that of \texttt{CIFAR-10-C}) to its test set  creating challenging scenarios for evaluating image recognition models. This dataset is commonly used in computer vision tasks to assess the performance of different models under distribution shifts.

\vspace{-.1in}
\subsection{Results}
\vspace{-.1in}
We estimate and compare ETAGE with some state-of-the-art methods in the literature namely, DeYo, EATA, SAR, and TENT. The results presented in~\cref{tab: cifar10} and~\cref{tab: cifar100} demonstrate that ETAGE consistently outperforms the other methods in the literature across a variety of corruption types. Specifically, on the CIFAR-10-C dataset (\cref{tab: cifar10}), ETAGE achieves the highest average accuracy, surpassing all other models. Similarly, on the CIFAR-100-C dataset (\cref{tab: cifar100}), ETAGE also leads with the highest average performance, indicating its robustness and effectiveness in handling diverse corruptions. These results validate the superiority of our approach, reinforcing ETAGE's capability to adapt effectively under varying conditions compared to state-of-the-art methods. It is also worth nothing that for TTA, lowering the batch size is challenging (lower batch sizes such as $1,2,4$ are known as wild scenarios). This is because TTA relies on calculating statistics  (mean and variance) over normalization layers (BN, GN, LN). When the batch size is small, these statistics may be less representative of the data distribution, leading to instability in the model's performance. The performance of different methods for different batch sizes are shown in~\cref{fig:batch}(b) for Gaussian noise corruption with severity $5$ of \texttt{CIFAR-10-C}. ETAGE retains its accuracy even with low batch sizes, while its counterparts' performance drop severely as batch size decreases. This underscores the effect of identifying and filtering out the noisy gradient samples which has been described in more details in~\cref{sec:method}. 

\cref{tab:metrics} provides a comprehensive comparison of different methods under Gaussian noise on \texttt{CIFAR-10-C}, including TENT, DEYO, SAR, EATA, and ETAGE, evaluated across key performance metrics: Expected Calibration Error (ECE), Maximum Calibration Error (MCE), Brier score, and Area Under the Receiver Operating Characteristic curve (AUROC). Lower ECE and MCE values suggest that a model's probability estimates are more dependable. Similarly, a lower Brier score indicates higher accuracy in probability predictions, while a higher AUROC value reflects better discrimination between classes. Notably, ETAGE outperforms others across most metrics, showcasing superior calibration and discriminative capability which in turn underscores the effectiveness of ETAGE in handling Gaussian noise while maintaining reliable predictions.

\vspace{-2mm}
\section{Conclusion} \label{sec:conclusiom}
\vspace{-2mm}

This study introduce \method, an improved method for TTA, addressing the limitations of existing approaches that primarily rely on entropy as a confidence metric. By integrating gradient norms with the PLPD, our approach effectively filters out noisy samples, leading to more stable and reliable model adaptation. The application of this method to the \texttt{CIFAR-10-C} and \texttt{CIFAR-100-C} datasets demonstrated its effectiveness in handling various distribution shifts, with consistent performance improvements over baseline methods. Our findings emphasize the importance of considering gradient information alongside entropy in TTA, providing a pathway to more robust model adaptation. As part of future work, we plan to extend our method to wild setups involving lower batch sizes and evaluate its performance on the \texttt{IMAGENET-C} dataset. These efforts will further assess the scalability and generalization of our approach in more diverse and challenging environments.
\bibliographystyle{IEEEbib}
\bibliography{__main}

\end{document}